\documentclass[10pt,twocolumn,letterpaper]{article}
\usepackage{iccv}

\usepackage{times}
\usepackage{epsfig}
\usepackage{graphicx}
\usepackage{amsmath,amssymb}
\usepackage{booktabs}
\usepackage{multirow}
\usepackage{pgf}
\usepackage{pgfplots}
\pgfplotsset{compat=1.18}
\usepackage{tikz}
\usetikzlibrary{
  positioning,
  matrix,
  fit,
  shapes,
  arrows.meta,
  shapes.multipart,
  calc,
  backgrounds
}

\usepackage[pagebackref=true,breaklinks=true,colorlinks,bookmarks=false]{hyperref}

\usepackage{svg}

\begin{document}

\title{Hybrid Cross-Device Localization via Neural Metric Learning and Feature Fusion}

\author{
Meixia Lin$^{*}$ \quad  Mingkai Liu$^{*}$ \quad Shuxue Peng \quad DiKai Fan \\
Shengyu Gu \quad Xianliang Huang \quad Haoyang Ye \quad Xiao Liu \\
PICO, ByteDance Inc. \\
{\tt\small \{meixia.lin, mingkai.liu, shuxue.peng, dikai.fan, shengyu.gu,}\\
{\tt\small xianliang.huang, haoyang.ye, xiao.liu\}@bytedance.com}
}

\maketitle
\thispagestyle{empty}

\let\thefootnote\relax\footnotetext{
$^{*}$Equal contribution. Meixia Lin led the project, designed the overall pipeline,
and conducted the neural map experiments and analysis. 
Mingkai Liu implemented the classical and retrieval branches, ran large-scale experiments,
and contributed to result validation. 
All authors discussed the results and contributed to the final report.
}

\begin{abstract}
We present a hybrid cross-device localization pipeline developed for the CroCoDL 2025 Challenge.
Our approach integrates a shared retrieval encoder and two complementary localization branches:
a classical geometric branch using feature fusion and PnP, and a neural feed-forward branch (MapAnything) for metric localization conditioned on geometric inputs.
A neural-guided candidate pruning strategy further filters unreliable map frames based on translation consistency, while depth-conditioned localization refines metric scale and translation precision on Spot scenes. 
These components jointly lead to significant improvements in recall and accuracy across both HYDRO and SUCCU benchmarks.
Our method achieved a final score of 92.62 (R@0.5m, 5°) during the challenge.
\end{abstract}

\begin{figure*}[t]
    \centering
    \includegraphics[width=0.8\textwidth]{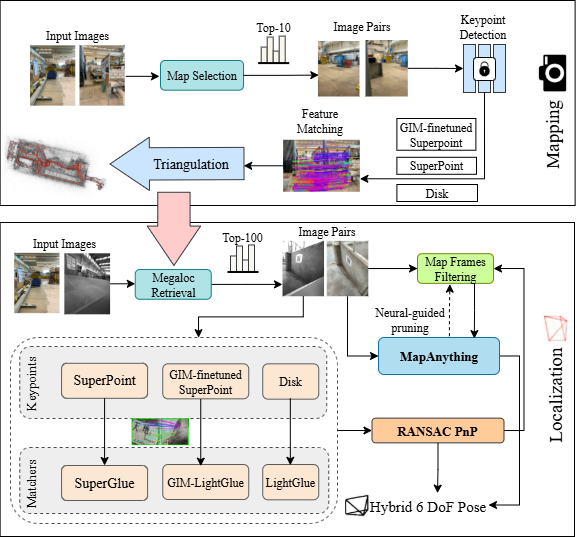}
    \caption{
    \textbf{Hybrid localization pipeline.}
    Our system consists of two stages: \textbf{mapping} (top) and \textbf{localization} (bottom).
    During mapping, input images are processed by \textit{keypoint detection} (SuperPoint, GIM-finetuned SuperPoint, DISK) 
    and \textit{feature matching} modules to construct the 3D map via triangulation.
    In the localization stage, \textit{MegaLoc retrieval} selects the top-100 candidate map frames for each query from the constructed map.
    The retrieved pairs are then processed by the \textit{classical branch} (feature matching + RANSAC-PnP) 
    and the \textit{neural branch} (MapAnything).
    A \textit{neural-guided map filtering} step prunes candidate frames with large translation distances before re-localization.
    Finally, a \textbf{hybrid 6-DoF pose} is produced by fusing PnP and neural predictions, achieving robust and accurate cross-device localization.
    }
    \label{fig:architecture}
\end{figure*}

\section{Method Overview}
\subsection{Unified Retrieval Encoder and Dual Localization Branches}
Both branches in our system share a single retrieval encoder to leverage the same high-quality candidate set and avoid redundant retrieval computations.
We employ MegaLoc~\cite{berton2025megaloc} to retrieve a set of top-$k$ map frames from the global database, providing the foundation for both localization paradigms.
In the classical geometric verification branch, the retrieved candidates are used for feature-based localization.
Each query–map pair undergoes multi-descriptor fusion (SuperPoint~\cite{deToro2018superpoint}, GIM-finetuned SuperPoint~\cite{shen2024gim}, DISK~\cite{tyszkiewicz2020disk}) and multi-matcher fusion (SuperGlue~\cite{schonberger2020superglue}, LightGlue~\cite{lindenberger2023lightglue}, GIM-LightGlue~\cite{shen2024gim}).
The aggregated matches are passed to COLMAP PnP estimation to compute a precise camera pose when sufficient inliers are available.
In the neural metric localization branch, the same retrieved candidates are used as geometric context inputs for MapAnything~\cite{keetha2025mapanything}.
Before re-localization, candidate frames are filtered by pose distance to the query ($\leq$20~m translation threshold) to remove irrelevant large-baseline views.
The remaining candidates, along with their depth maps, intrinsics, and poses, are fed into MapAnything~\cite{keetha2025mapanything} for feed-forward metric re-localization.

\subsection{Feature Fusion Retrieval Pipeline}
To improve retrieval robustness under cross-device and cross-domain conditions, we design a feature fusion retrieval pipeline that aggregates complementary descriptors and matchers before localization.
The retrieval backbone (MegaLoc~\cite{berton2025megaloc}) provides global candidates, while our fusion mechanism enhances local discriminability and resilience to viewpoint or illumination changes.
Specifically, we combine multiple descriptors — SuperPoint~\cite{deToro2018superpoint}, GIM-finetuned SuperPoint~\cite{shen2024gim}, and DISK~\cite{tyszkiewicz2020disk} — to capture both low-level geometric structure and high-level semantic cues.
For matching, we integrate SuperGlue~\cite{schonberger2020superglue}, LightGlue~\cite{lindenberger2023lightglue}, and GIM-LightGlue~\cite{shen2024gim}, each contributing distinct correspondence priors.
Matches are aggregated using a confidence-weighted union, and duplicate correspondences are pruned based on spatial consistency.
This fusion pipeline leads to stronger recall and higher matching reliability across heterogeneous camera domains.
It also improves downstream PnP estimation stability, serving as a bridge between global retrieval and fine-grained pose estimation.

\subsection{Hybrid Pose Estimation (PnP + MapAnything)}
In practice, PnP may fail for cross-device cases due to insufficient feature correspondences or calibration noise.
We therefore design a hybrid pose estimation scheme:
when PnP succeeds with a sufficient number of inliers ($>$120), the query pose from the PPL pipeline ($T_q^{PnP}$) is used; 
otherwise, the pose is predicted by MapAnything ($T_q^{MA}$)~\cite{keetha2025mapanything}.
This hybrid strategy guarantees coverage while maintaining geometric consistency.
The estimated pose ($\hat{T}_q$) is then used to guide map candidate filtering.

\subsection{Neural-guided Candidate Pruning}
Given the retrieved map frames $\{I_i, T_i = [R_i|t_i]\}$ with known accurate poses, and the query pose $\hat{T}_q$ estimated from either PnP or MapAnything, we perform a neural-guided pruning stage to discard geometrically inconsistent maps. We compute translation distances $d_i = \|\hat{t}_q - t_i\|_2$ and retain a subset $S = \{i \mid d_i \leq 20\,\text{m}\}$. Localization is then re-run on the pruned set $S$ using MapAnything~\cite{keetha2025mapanything}. This hybrid strategy enhances robustness in large-scale scenes, where excessive translation gaps between the query and map frames may degrade feed-forward predictions. It can be regarded as a \textit{neural-assisted geometric verification}: neural predictions guide candidate selection, while geometric constraints ensure physical consistency.
\vspace*{-2mm}
\subsection{Depth Conditioning and Filtering}
For datasets that provide depth information (e.g., Spot), we further incorporate depth maps to enhance metric scale estimation within MapAnything~\cite{keetha2025mapanything}.
Each depth map is transformed into the query camera coordinate system using the known extrinsic parameters before being fed into the network.
This depth conditioning significantly improves the accuracy of translation and absolute scale, reducing average translation error by a large margin.
However, when depth maps contain noise or inaccurate measurements, they can introduce instability in rotation estimation.
To mitigate this, we perform depth filtering by removing pixels with invalid or outlier depth values before feeding them into MapAnything~\cite{keetha2025mapanything}.
This step ensures that only reliable geometric cues contribute to the neural relocalization process, balancing scale precision and rotational stability across diverse scenes.

\begin{table}[htbp]
  \centering
  \caption{HYDRO}
  \label{tab:hydro}
  \begin{tabular}{|c|c|c|c|c|}
    \hline
    \multicolumn{2}{|c|}{} & \multicolumn{3}{c|}{map} \\
    \cline{3-5}
    \multicolumn{2}{|c|}{} & iOS & HL & Spot \\
    \hline
    \multirow{3}{*}{\centering query} & iOS & 95.82\% & 98.53\% & 88.70\% \\
    \cline{2-5}
    & HL & 93.87\% & 98.69\% & 95.40\% \\
    \cline{2-5}
    & Spot & 90.69\% & 97.39\% & 98.88\% \\
    \hline
  \end{tabular}
\end{table}

\begin{table}[htbp]
  \centering
  \caption{SUCCU}
  \label{tab:succu}
  \begin{tabular}{|c|c|c|c|c|}
    \hline
    \multicolumn{2}{|c|}{} & \multicolumn{3}{c|}{map} \\
    \cline{3-5}
    \multicolumn{2}{|c|}{} & iOS & HL & Spot \\
    \hline
    \multirow{3}{*}{\centering query} & iOS & 86.38\% & 87.70\% & 80.56\% \\
    \cline{2-5}
    & HL & 94.47\% & 96.64\% & 79.91\% \\
    \cline{2-5}
    & Spot & 92.93\% & 90.57\% & 100.00\% \\
    \hline
  \end{tabular}
\end{table}

\begin{table}[htbp]
  \centering
  \caption{OVERALL}
  \label{tab:overall}
  \begin{tabular}{|c|c|c|c|c|}
    \hline
    \multicolumn{2}{|c|}{} & \multicolumn{3}{c|}{map} \\
    \cline{3-5}
    \multicolumn{2}{|c|}{} & iOS & HL & Spot \\
    \hline
    \multirow{3}{*}{\centering query} & iOS & 91.10\% & 93.11\% & 84.63\% \\
    \cline{2-5}
    & HL & 94.17\% & 97.66\% & 87.66\% \\
    \cline{2-5}
    & Spot & 91.81\% & 93.98\% & 99.44\% \\
    \hline
  \end{tabular}
\end{table}

\section{Results and Discussion}
Our method achieved a final score of 92.62 on the HYDRO and SUCCU datasets (R@0.5m). Multi-descriptor fusion improved recall by 3 percentage points. Beyond quantitative metrics, a key observation lies in the robustness of MapAnything~\cite{keetha2025mapanything} under unseen or cross-device conditions. When the query frame exhibits little or no visual overlap with mapped regions, traditional PPL pipelines, which rely on explicit feature correspondences and PnP estimation, often fail to recover a stable camera pose. In contrast, MapAnything~\cite{keetha2025mapanything} maintains geometric robustness by leveraging geometric priors and multi-view consistency, inferring plausible camera poses even when local feature matches are sparse or noisy. This generalization ability to unseen scenes complements the high precision of classical geometric reasoning, leading to a more reliable cross-device localization framework overall.

\section{Conclusion}
We introduced a hybrid cross-device localization framework combining traditional geometric pipelines and feed-forward neural reconstruction.
By coupling MapAnything-based pose estimation with PnP-based localization and neural-guided pruning, our system achieves strong robustness, precision, and scalability across diverse devices and scenes.
The design demonstrates that neural metric geometry can complement classical geometric reasoning, bridging the gap between structure-based and feed-forward localization paradigms.

{
    \small
    \bibliographystyle{ieeenat_fullname}
    \bibliography{main}

@String(CVPR= {IEEE Conf. Comput. Vis. Pattern Recog.})

@String(CVPRW= {IEEE Conf. Comput. Vis. Pattern Recog. Worksh.})

@String(CVPR  = {CVPR})

@String(CVPRW= {CVPRW})

@article{keetha2025mapanything,
  title={MapAnything: Universal Feed-Forward Metric 3D Reconstruction},
  author={Keetha, Nikhil and M{\"u}ller, Norman and Sch{\"o}nberger, Johannes and Porzi, Lorenzo and Zhang, Yuchen and Fischer, Tobias and Knapitsch, Arno and Zauss, Duncan and Weber, Ethan and Antunes, Nelson and others},
  journal={arXiv preprint arXiv:2509.13414},
  year={2025}
}

@article{shen2024gim,
  title={Gim: Learning generalizable image matcher from internet videos},
  author={Shen, Xuelun and Cai, Zhipeng and Yin, Wei and M{\"u}ller, Matthias and Li, Zijun and Wang, Kaixuan and Chen, Xiaozhi and Wang, Cheng},
  journal={arXiv preprint arXiv:2402.11095},
  year={2024}
}

@inproceedings{deToro2018superpoint,
  title={SuperPoint: Self-Supervised Interest Point Detection and Description},
  author={DeTone, Daniel and Malisiewicz, Tomasz and Rabinovich, Andrew},
  booktitle={Proceedings of the IEEE Conference on Computer Vision and Pattern Recognition Workshops (CVPRW)},
  year={2018}
}

@inproceedings{schonberger2020superglue,
  title={SuperGlue: Learning Feature Matching with Graph Neural Networks},
  author={Sch{\"o}nberger, Johannes L and Frahm, Jan-Michael},
  booktitle={Proceedings of the IEEE/CVF Conference on Computer Vision and Pattern Recognition (CVPR)},
  pages={4938--4947},
  year={2020}
}

@inproceedings{lindenberger2023lightglue,
  title={Lightglue: Local feature matching at light speed},
  author={Lindenberger, Philipp and Sarlin, Paul-Edouard and Pollefeys, Marc},
  booktitle={Proceedings of the IEEE/CVF international conference on computer vision},
  pages={17627--17638},
  year={2023}
}

@article{tyszkiewicz2020disk,
  title={Disk: Learning local features with policy gradient},
  author={Tyszkiewicz, Micha{\l} and Fua, Pascal and Trulls, Eduard},
  journal={Advances in neural information processing systems},
  volume={33},
  pages={14254--14265},
  year={2020}
}

@inproceedings{berton2025megaloc,
  title={Megaloc: One retrieval to place them all},
  author={Berton, Gabriele and Masone, Carlo},
  booktitle={Proceedings of the Computer Vision and Pattern Recognition Conference},
  pages={2861--2867},
  year={2025}
}
}

\end{document}